\documentclass[letterpaper]{article} 
\usepackage{aaai24}  
\usepackage{times}  
\usepackage{helvet}  
\usepackage{courier}  
\usepackage[hyphens]{url}  
\usepackage{graphicx} 
\usepackage{natbib}  
\usepackage{caption} 
\usepackage{algorithm}
\usepackage{algorithmic}

\usepackage{newfloat}
\usepackage{listings}
\DeclareCaptionStyle{ruled}{labelfont=normalfont,labelsep=colon,strut=off} 
\floatstyle{ruled}
\newfloat{listing}{tb}{lst}{}
\floatname{listing}{Listing}
\usepackage{makecell}
\usepackage{multirow} 
\usepackage{multicol} 
\usepackage{arydshln}
\usepackage{amssymb}
\usepackage{amsmath}
\usepackage{booktabs}
\usepackage{subfigure}
\usepackage{color}
\usepackage{xcolor}

\newcommand{\ie}{\textit{i}.\textit{e}.}

\newcommand{\etal}{\textit{et}~\textit{al}.}

\title{HR-Pro: Point-supervised Temporal Action Localization \\ via Hierarchical Reliability Propagation }
\newcommand{\modelname}{HR-Pro}

\author{
    Huaxin Zhang\textsuperscript{\rm 1},
    Xiang Wang\textsuperscript{\rm 1},
    Xiaohao Xu\textsuperscript{\rm 2},
    Zhiwu Qing\textsuperscript{\rm 1},
    Changxin Gao\textsuperscript{\rm 1},
    Nong Sang\textsuperscript{\rm 1}\thanks{indicates corresponding author.}
}

\affiliations{
        \textsuperscript{\rm 1} Key Laboratory of Image Processing and Intelligent Control,\\
    School of Artificial Intelligence and Automation, Huazhong University of Science and Technology \\
        \textsuperscript{\rm 2} University of Michigan, Ann Arbor\\
        \{zhanghuaxin, wxiang, qzw, cgao, nsang\}@hust.edu.cn,
        \{xiaohaox\}@umich.edu

}

\begin{document}

\maketitle
\begin{abstract}

Point-supervised Temporal Action Localization (PSTAL) is an emerging research direction for label-efficient learning.
However, current methods mainly focus on optimizing the network either at the snippet-level or the instance-level,
neglecting the inherent reliability of point annotations at both levels.
In this paper, we propose a \textbf{H}ierarchical \textbf{R}eliability \textbf{Pro}pagation (\textbf{\modelname})
framework, which consists of two reliability-aware stages: Snippet-level Discrimination Learning and Instance-level Completeness Learning, both stages explore the efficient propagation of high-confidence cues in point annotations.
For snippet-level learning, we introduce an online-updated memory to store reliable snippet prototypes for each class. 
We then employ a Reliability-aware Attention Block to capture both intra-video and inter-video dependencies of snippets, resulting in more discriminative and robust snippet representation.
For instance-level learning, we propose a point-based proposal generation approach as a means of connecting snippets and instances, which produces high-confidence proposals for further optimization at the instance level.
Through multi-level reliability-aware learning, we obtain more reliable confidence scores and more accurate temporal boundaries of predicted proposals.
Our \modelname~achieves state-of-the-art performance on multiple challenging benchmarks, including an impressive average mAP of \textit{\textbf{60.3\%}} on THUMOS14. Notably, our \modelname~largely {surpasses all previous point-supervised methods, and even outperforms several competitive fully-supervised methods}.
Code will be available at https://github.com/pipixin321/HR-Pro.

\end{abstract}

\section{Introduction}
\begin{figure}[t]
\centering
\includegraphics[scale=0.5]{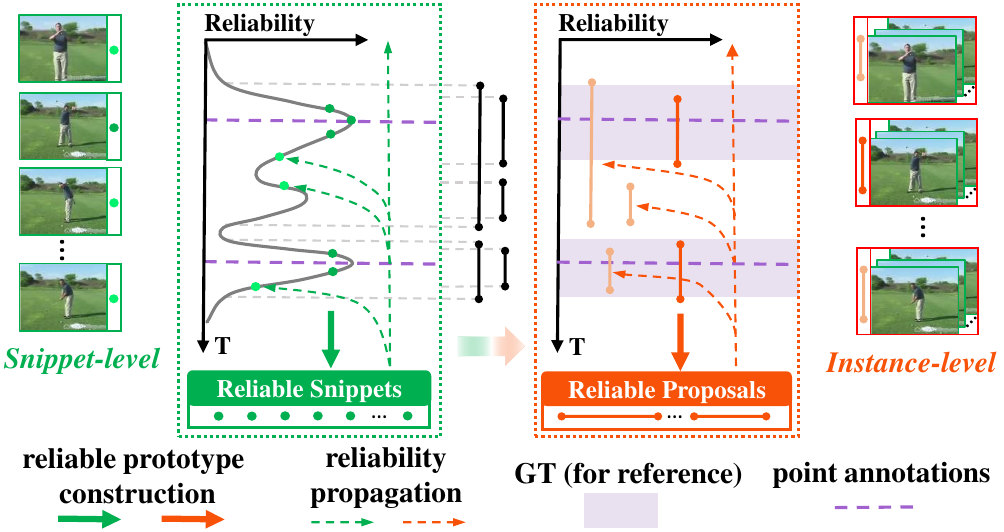}
\caption{Motivation illustration. Given the point-level annotation (in purple), we consider the action reliability prior both at the snippet level and instance proposal level to enable reliability-aware action representation learning. 
In specific, our insight is to propagate reliable prototypes to produce more discriminative snippet-level scores and more reliable and complete instance-level scores.
Darker color (greener or more orange) indicates higher reliability. Here, a case with one action class is shown for brevity.
}
\label{teaser}
\end{figure}

Temporal action localization is a fundamental task in video understanding field, which attempts to temporally localize and classify action instances in the untrimmed video, and has attracted increasing attention due to its potential application in various fields~\cite{lee2012discovering,vishwakarma2013survey}.
However, traditional fully-supervised methods~\cite{BSN,BMN,GTAD,qing2021temporal,wang2022context,wang2022rcl,nag2022proposal} require accurate temporal annotations, which are extremely time-consuming and labor-demanding, hindering the practical applications.
Therefore, many researchers~\cite{wang2017untrimmednets,shou2018autoloc,wang2021weakly} start to pay attention to weakly-supervised temporal action localization (WSTAL) where only video-level labels are available.
Although significant progress in WSTAL has been made, the lack of action boundary information imposes a great challenge for models to distinguish actions and backgrounds, resulting in unsatisfactory performance compared to fully-supervised methods.

Under the WSTAL setting, to balance labeling cost and detection performance, Ma \etal~\cite{ma2020sf} introduce the point-supervised temporal action localization (PSTAL) task which provides only a timestamp label for each action instance. 
Their pioneering research indicates that point-level annotations consume almost comparable labor costs as video-level annotations while providing richer guidance information.
Subsequently, many works start to follow this setting and propose various customized solutions.
Typically, LACP~\cite{lee2021learning} proposes to learn completeness of action instances by searching the optimal pseudo sequence with a greedy algorithm.
Ju~\etal~\cite{ju2021divide} proposes a seed frame detector to generate proposals and then performs regression and classification on the proposals.

Although previous methods have achieved impressive results, they are still limited to optimizing the network at either the snippet level or instance level.
Snippet-level approaches~\cite{ma2020sf,lee2021learning} may produce many unreliable (\textit{e.g.}, overcomplete or false positive) detections because they only consider individual snippets, which ignore the overall action instance. 
On the other hand, instance-level approach~\cite{ju2021divide} cannot achieve satisfactory optimization due to the absence of reliable proposals generated from snippet scores.
We propose that the high reliability of point annotations can be propagated at both snippet and instance levels.
To accomplish this, we derive reliable prototypes at different levels by considering their confidence scores and relative positions to point annotations.
Leveraging these reliable prototypes through high-confidence information propagation enables the network to learn more discriminative snippet-level representation and more reliable instance-level proposals.

Building upon these insights, we present a Hierarchical Reliability Propagation method that consists of two reliability-aware stages: Snippet-level Action Discrimination Learning and Instance-level Action Completeness Learning. These stages are illustrated in Fig.~\ref{teaser}.
(1) In the Snippet-level Action Discrimination Learning stage, our objective is to obtain discriminative snippet-level scores for generating more reliable proposals. To achieve this, we introduce an online-updated memory to store reliable prototypes for each class. Additionally, we propose a Reliability-aware Attention Block to propagate high-confidence cues from these reliable prototypes to other snippets. Through contrastive optimization of the memory and snippet features, we derive a more discriminative action representation.
(2) In the Instance-level Action Completeness Learning stage, we refine the confidence score and boundary of the proposals through instance-level feature learning. We propose a point-based proposal generation method that produces reliable instance-level prototype proposals, along with high-confidence positive and negative proposals. These proposals' features are then fed
into a Score Head and a Regression Head to predict the completeness score and refined boundary. This prediction process is guided by reliable instance prototypes. As a result, the network can estimate more reliable instance-level scores and achieve more accurate temporal boundaries.

To summarize, our contributions are as follows: 
\begin{itemize}
\item Our proposed method, \textit{i.e.}, {\modelname}, 
is the first to leverage inherent reliability of point annotations at both snippet and instance level optimization in the PSTAL domain.

\item At the snippet level, we propose a reliability-aware attention module and reliable-memory-based contrastive loss to acquire discriminative snippet-level representation. 

\item At the instance level, we propose reliability-based proposal generation and ranking method to produces high-confidence proposals for further optimization at the instance level.

\item 
Our \modelname~achieves state-of-the-art performance on four standard temporal action localization benchmarks, including an impressive average mAP of \textit{\textbf{60.3\%}} on THUMOS14, \textbf{which even surpasses several competitive fully-supervised methods}.
\end{itemize}

\section{Related Work}

\textbf{Fully-supervised temporal action localization.} 
{Mainstream fully-supervised methods can be divided into two categories, \ie, one-stage and two-stage.}
The one-stage methods~\cite{GTAD,zhang2022actionformer} simultaneously predicts the boundary and category of action as the final detection result. 
The two-stage methods~\cite{BSN,BMN,qing2021temporal,wang2022context,wang2021proposal} first generate numerous proposals and then classify the proposals.
Despite the significant progress in recent years, these fully-supervised methods require expensive annotation costs, which limits their application.

\noindent\textbf{Weakly-supervised temporal action localization.} 
{To reduce the labeling cost, many weakly-supervised temporal action localization (WSTAL) methods ~\cite{wang2017untrimmednets,shou2018autoloc,liu2019weakly} have been proposed where only video-level labels are available.} 
Most recent WSTAL methods follow the localization-by-classification mode.
They first use a snippets classifier to evaluate the class probability of each video snippet, \ie, Class Activation Sequence (CAS), and then locate the temporal boundary using multiple predefined thresholds.
{Recently, many attempts have been made to enhance the performance of the model.}
BaS-Net~\cite{lee2020background} introduces background classes and background branch to suppress class activation values of background snippets. 
ACM-Net~\cite{qu2021acm} proposes three attention branches to separate foreground, background, and context. 
CoLA~\cite{zhang2021cola} proposes a hard snippet mining algorithm and a snippet contrastive loss to refine the hard snippet representation in feature space.
ACG-Net~\cite{yang2022acgnet} and DGCNN~\cite{shi2022dynamic} adopt graph networks to enhance feature embedding and model relationships between action snippets.
ASM-Loc~\cite{he2022asm} proposes to use intra- and inter-segment attention for modeling action dynamics and capturing temporal dependencies.
{Due to the absence of frame-wise annotations, the perfomance of these models falls largely behind the fully-supervised methods.}

\noindent\textbf{Point-supervised temporal action localization.}
To balance labeling cost and model performance, point-supervised temporal action localization (PSTAL) task is proposed by~\cite{ma2020sf}, which provides a timestamp label for each action instance. 
To explore the guidance information provided by point annotations,
SF-Net~\cite{ma2020sf} uses the single-frame label to mine its adjacent pseudo label for training classifiers. 
Ju et.al~\cite{ju2021divide} uses a two-stage approach, which proposes a seed frame detector to generate proposals and then performs regression and classification on the proposals.
LACP~\cite{lee2021learning} searches the optimal pseudo sequence through a greedy algorithm which is used to guide the network to learn the completeness of action instances. 
However, these methods are limited in optimizing the network either at snippet-level or at instance-level, leading to less effective discriminative representations at the snippet-level and unreliable scores at the instance-level.

\begin{figure*}[t]
    \centering
    \includegraphics[width=\textwidth]{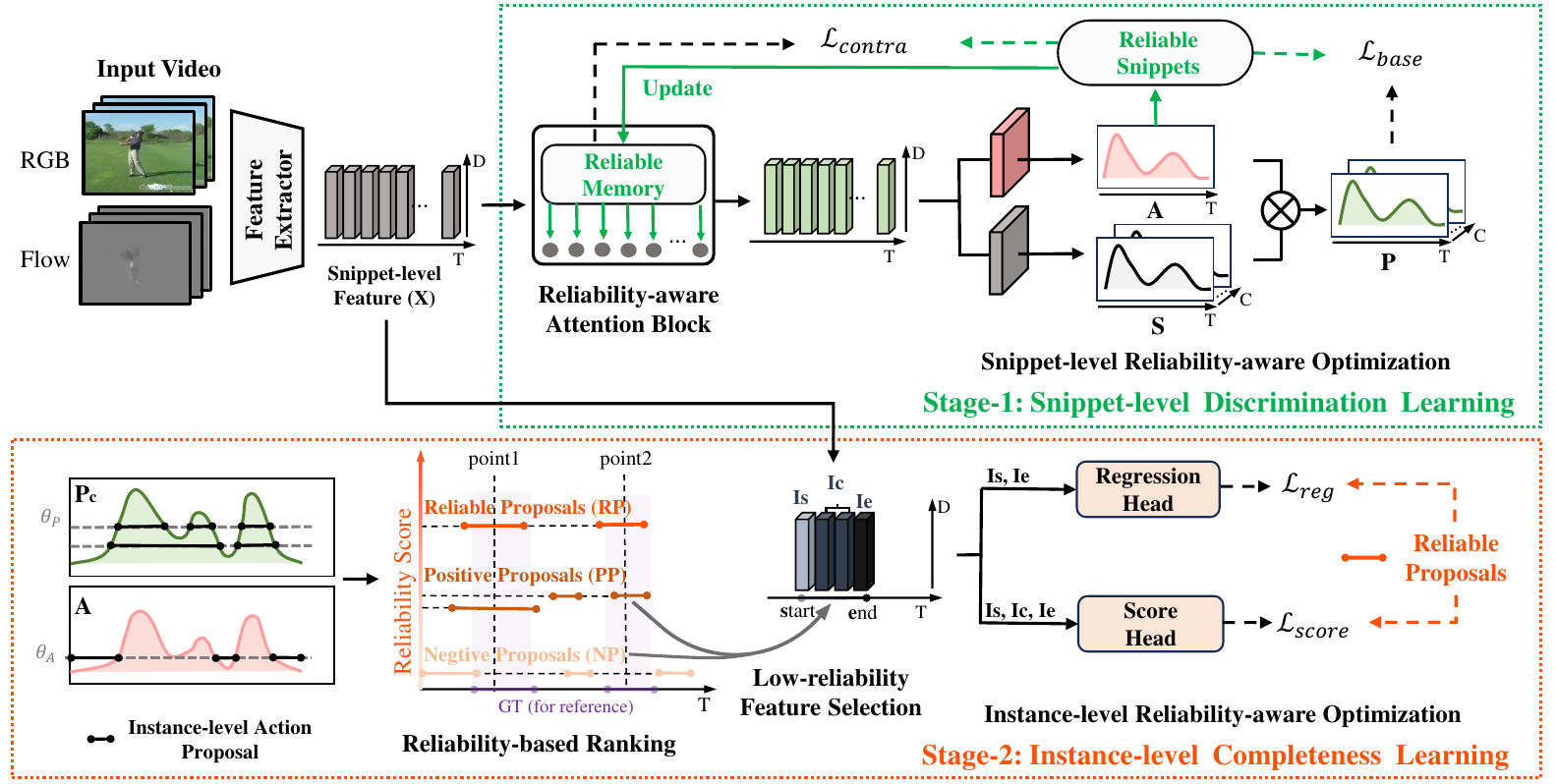}
    \caption{
Overview of Hierarchical Reliability Propagation (\modelname).
We propagate reliable prototypes during two-stage action localization learning, \textit{i.e.}, Snippet-level Discrimination Learning and Instance-level Completeness Learning.
(1) \textit{Snippet} level: we aim to obtain snippet representations with good inter-class discrimination and action-background discrimination.
(2) \textit{Instance} level: we aim to refine the confidence
score and boundary of the coarse proposals generated from snippet-level output.
}
\label{model_architecture}
\end{figure*}
\section{Preliminaries}

\noindent{\textbf{Problem Definition}.} For point-supervised temporal action localization (PSTAL), models are trained with a single-frame annotation for each untrimmed video. Each action instance is annotated with a temporal point $p_i$ and a one-hot vector $y_{p_i}$ indicating the action category $c$ with $y_{p_i}[c]=1$. The video contains a total of $N$ action instances.
During inference, we generate predicted results for each test video using ${(s_m, e_m, c_m, p_m)}_m^M$, where $s_m$ and $e_m$ are the start and end times of the $m$-th predicted action instance, $c_m$ is the predicted category, and $p_m$ is the confidence score. $M$ is the total number of predicted action instances.

\noindent\textbf{Baseline Architecture.}
The input video is first divided into multi-frame snippets, then we use a pre-trained video classification model to extract RGB and optical flow features of each snippet and concatenate them along the channel dimension. The  features of the input video are formulated as $\mathbf{X}\in{\mathbb R}^{T\times D}$, where $T$ and $D$ indicate the number of snippets and the dimension of features, respectively. The features are then fed into a feature embedding layer get task-specific embedded features $\mathbf{X}^e\in{\mathbb R}^{T\times D}$. 

Following previous work~\cite{lee2021learning}, we first input the embedded features into a snippet-level classifier to obtain class-specific activation sequence (CAS) $\mathbf{S}\in{\mathbb R}^{T\times C}$, where $C$ denotes the class number. To reduce the noise from background snippets, we use a convolutional layer to generate a class-agnostic attention sequence $\mathbf{A}\in{\mathbb R}^T$. 
Then, we fuse them by element-wise production to get the final snippet-level predictions $\mathbf{P}\in{\mathbb R}^{T\times C}$ where $\mathbf{P}=\mathbf{S}\cdot \mathbf{A}$.

\noindent\textbf{Baseline Optimization Loss.} Based on the characteristic that each action instance contains a point annotation and the adjacent point annotations are in different action instances, we select pseudo action snippets $\mathcal{T}^+=\{t_i\}_{i=1}^{N_{{act}}}$ and background snippets $\mathcal{T}^-=\{t_j\}_{j=1}^{N_{bkg}}$ based on point annotations and class-agnostic attention sequence.
Specifically, snippets near a point annotation with higher class-agnostic attention than a given threshold are labeled as pseudo-action snippets, which share the same action category as the point annotation. Conversely, snippets located between two adjacent point annotations with the lowest class-agnostic attention or lower class-agnostic attention than a given threshold are labeled as pseudo-background snippets.
We use these pseudo snippet samples for supervision:
\begin{equation}
\begin{small}
{\mathcal L}_{base}=\frac{1}{N_{act}} \sum_{c=1}^{C} \sum_{t\in{\mathcal{T}^+}}^{N_{act}}FL(\mathbf{P}_{t,c})
 + \frac{1}{N_{bkg}} \sum_{t\in{\mathcal{T}^-}}^{N_{bkg}} FL(1-\mathbf{A}_{t})
\label{eq_4}
\end{small}
\end{equation}
where $N_{act}$ and $N_{bkg}$ is the total number of pseudo action snippets and background snippets respectively, \textit{FL} represents the focal loss function~\cite{focalloss}.

\section{Method: Hierarchical Reliability Propagation}

Reliability can help the network mine more pseudo samples, which can alleviate the sparsity of guidance in point-supervised setting.
We argue that the inherent reliability of point annotations can be propagated at both snippet and instance level optimization.
Therefore, we propose a Hierarchical Reliability Propagation framework, which divides action localization learning into two cascaded stages: (1) Snippet-level Action Discrimination Learning and (2) Instance-level Action Completeness Learning.

\subsection{\textbf{Snippet-level Action Discrimination Learning}}

Previous works have primarily focused on estimating temporal pseudo-labels to expand training samples,
which restricts the propagation of high-confidence snippet information within a single video.
Thus, we introduce Reliability-aware Snippet-level Action Discrimination Learning, which proposes to store the reliable prototypes for each class and propagate high-confidence cues from these prototypes to other snippets via intra-video and inter-video ways.

\noindent\textbf{Reliable Prototype Construction.}
As the snippet-level action representation, \textit{i.e.}, snippet features, only captures short-term and partial action states, the feature can be noisy and unreliable. Thus, our insight is to construct reliable snippet prototypes via a de-noising mechanism for further reliability-guided optimization. 

Specifically, we create an online-updated prototype memory to store reliable prototypes for each class during representation learning, enabling us to leverage the feature information from the entire dataset to mitigate the noise of each feature. Formally, we denote the item in memory by $\mathbf{m_c} \in \mathbb{R}^ D(c =1,2,...,C)$.
Under the PSTAL setting, we initialize the prototype pool by selecting features with point annotations for each class. This is done by computing the average of snippet features $x_{p_i}$ corresponding to the point annotations $p_i$ for class $c$. We normalize the sum by the total number of point annotations $N_c$ for class $c$ across all training videos. The initial prototype memory is defined as:
\begin{equation}
\mathbf{m}_c^0=\frac{1}{N_c}\sum_{i}^{N_c}{x_{p_i}}(y_{p_i}[c]=1)
\end{equation}
Next, we update the prototypes for each class using the features of pseudo-action snippets, which is formulated as:
\begin{equation}
\mathbf{m}_{c}^{t}=\mu \mathbf{m}_{c}^{(t-1)}+(1-\mu)\mathbf{x}_{t_i}^{(t)}
\end{equation}
Here, $\mu$ denotes the momentum coefficient for an update.

\begin{figure}[t]
\centering
\includegraphics[scale=0.41]{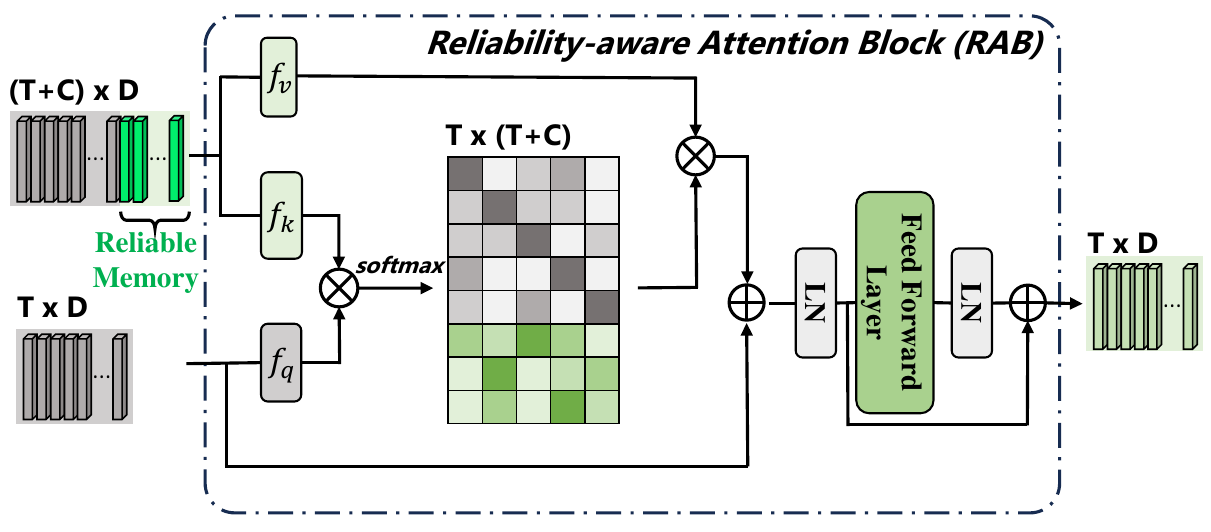}
\caption{Architecture detail of Reliability-aware Attention Block (RAB). Reliable prototype memory (in green) is injected into the original snippet features (in grey) to introduce reliable cues via the attention mechanism.}
\label{fig:RAB}
\end{figure}
As is shown in Fig.~\ref{fig:RAB}, to derive the prototype, we input the snippet-level features extracted by the feature extractor into a Reliability-aware Attention Block (RAB). 
The RAB is specifically designed to capture both intra-video and inter-video dependencies of snippets, enabling the modeling of complementary temporal relationships.
Long-term temporal dependency modeling is crucial for long videos, as supported by previous works~\cite{zhang2022actionformer,Wang_2022_CVPR,RPCMVOS,wang2023molo}. However, attention tends to become sparse and focus mainly on discriminative snippets within the same video, resulting in limited information interaction.
Therefore, the RAB incorporates the insight of propagating global class information from a reliable prototype (\textit{i.e.}, snippet) memory, thereby enhancing the robustness of snippet features and increasing the attention on less discriminative snippets.

Technically, we employ a linear layer $f_q$ to project the video features onto the corresponding query. Subsequently, we concatenate ($[;]$ denotes concatenation) the video features $\mathbf{X}$ with the prototype features $\mathbf{m}_i$ stored in the reliable memory bank. Then, we use separate linear layers, $f_k$ and $f_v$, to project the concatenated features into key and value, respectively:
\begin{equation}
\mathbf{Q}=f_q(\mathbf{X})
\end{equation}
\begin{equation}
\mathbf{K}=f_k({[\mathbf{X};\mathbf{m}_1;...;\mathbf{m}_C]}),
\mathbf{V}=f_v({[\mathbf{X};\mathbf{m}_1;...;\mathbf{m}_C]})
\end{equation}

Next, we multiply the query and the transposed key to obtain the non-local attention $\mathbf{attn}\in {\mathbb{R}}^{T \times (T + C)}$.
\begin{equation}
\textbf{attn}=softmax(\mathbf{Q} \cdot \mathbf{K}^T / \sqrt{D}) 
\end{equation}

Furthermore, we multiply the attention with the value, and pass it through a Feed Forward Layer (FFL) composed of cascaded FC-GELU-FC layers. Here, LayerNorm (LN) and residual connection are used for the normalization and retention of the original information.
The output reliability-aware features are fed into subsequent network layers for temporal action localization.

\noindent\textbf{Reliability-aware Optimization.} 
To push away the features of pseudo-action snippets and prototypes from different classes in the reliable prototype pool and push away the features of pseudo-action and background snippets in the same video, we follow a contrastive learning manner and propose a reliability-aware snippet-contrastive loss (${\mathcal L}_{contra}$):
\begin{equation}
\begin{small}
\begin{aligned}
{\mathcal L}_{contra} = -\frac{1}{C}\sum_{c=1}^{C}\sum_{t_i^c}log(
                        \frac{s(\mathbf{x}_{t_i^c},\mathbf{m}_c)}
                        {s(\mathbf{x}_{t_i^c},\mathbf{m}_c)+\sum_{\forall k \neq c}s(\mathbf{x}_{t_i^c},\mathbf{m}_k)} \\
                        +\frac{s(\mathbf{x}_{t_i^c},\mathbf{m}_c)}
                        {s(\mathbf{x}_{t_i^c},\mathbf{m}_c)+\sum_{\forall t_j \in \mathcal{T}^-}s(\mathbf{x}_{t_j},\mathbf{m}_c)})
\end{aligned}
\end{small}
\end{equation}
where $t_i^c$ indicates the pseudo action snippet of class c; $s(\cdot,\cdot)$ is the similarity function formulated as $s(\mathbf{x_1},\mathbf{x_2}) = exp(\bar{\mathbf{x}}_1 \cdot \bar{\mathbf{x}}_2 / \tau)$ with a temperature parameter $\tau$, $\bar{\mathbf{x}}$
represent the normalized features of $\mathbf{x}$,

Finally, the overall training objective for Snippet-level Action Discrimination Learning includes both the baseline loss ${\mathcal L}_{base}$ and our reliability-aware snippet optimization loss ${\mathcal L}_{contra}$ weighted by a parameter $\lambda_1$:
\begin{equation}
{\mathcal L}_{snippet} = {\mathcal L}_{base} + \lambda_1{\mathcal L}_{contra}
\end{equation}

\subsection{\textbf{Instance-level Action Completeness Learning}}

Snippet-level Representation Learning empowers our model with robust snippet-level action discrimination capabilities. However, a snippet-level-based pipeline can produce numerous unsatisfactory detections despite discriminative snippet representation, because the proposal score is unreliable without considering the whole instance
(\textit{e.g.}, \textit{running} in background frames has a high snippet score in the \textit{long jump} category, but it is not a complete \textit{long jump} action).
To fully explore the temporal structure of actions at the instance level and optimize the score ranking of proposals, we introduce Instance-level Action Completeness Learning.
This method is aimed at refining the proposals' confidence score and boundary via instance-level feature learning with the guidance of reliable instance prototypes.

\noindent\textbf{Reliable Prototype Construction.} 
To leverage instance-level priors of point annotations during training, we propose a point-based proposal generation method that yields reliable instance-level prototype proposals, along with high-confidence positive and negative proposals.
Initially, we produce candidate proposals for each predicted class by selecting snippets with class-specific activation scores higher than the threshold $\theta_{P}$. (we use multiple thresholds in implementation.)
The OIC (outer-inner-contrast) score~\cite{shou2018autoloc} is calculated for each candidate proposal to gauge its reliability score, {represented as $p_{OIC}$.
Lower reliability scores indicate incomplete or over-complete predictions.
We formulate each candidate proposal as $(s_i,e_i,c_i,p_{OIC})$,}
these proposals are then ranked into two types based on their reliability score and temporal position: (1) Reliable Proposals ($\mathbf{R}\mathbf{P}$): for each point in each class, the proposal contained this point, and has the highest reliability (i.e., OIC score); (2) Positive Proposals ($\mathbf{P}\mathbf{P}$): all remaining candidate proposals.
To ensure a balanced number of positive and negative samples, we group snippets with class-agnostic attention scores lower than pre-defined threshold $\theta_{A}$, which derives Negative Proposals ($\mathbf{N}\mathbf{P}$).

\noindent\textbf{Reliability-aware Optimization.} 
For each proposal, we select all snippet features within the proposal region as its center features $I_c$, then we expand the boundary of the proposal with ratio $\varepsilon$ to get starting region and ending region, which derives starting feature $I_s$ and ending features $I_e$ of the proposal, $\varepsilon$ is set to 0.25 in practice.

(1) To predict the completeness score of each proposal, we use the boundary-sensitive proposal features following~\cite{ren2023proposal} as input to the Score Head $\phi_s$.
\begin{equation}
\hat{p}_{comp}=\phi_{s}([\overline{I}_c-\overline{I}_s;\overline{I}_c;\overline{I}_c-\overline{I}_e])
\end{equation}
where $\overline{I}_s,\overline{I}_c,\overline{I}_e$ is the max-pooling feature of $I_s,I_c,$ and $I_e$ along the temporal dimension, respectively. 

Then, the reliability-aware supervision for instance level completeness score can be formulated as:
\begin{equation}
{\mathcal L}_{score}=\frac{1}{N_p+N_n} \sum_{i=1}^{N_p+N_n}Smooth{L1}(\hat{p}_{comp},g_{comp})
\end{equation}
where $N_p$, $N_n$ are the total number of positive proposals and negative proposals, respectively, $g_{comp}$ represents the Intersection over Union (IoU) between the proposal and the most reliable {proposal} ($\mathbf{R}\mathbf{P}$) that matches it.

(2) To obtain more accurate action proposal boundaries, we input the starting features and ending features of each proposal in $\mathbf{PP}$ into the Regression Head $\phi_{r}$ to predict the offset of the start time and the end time, \ie, $\Delta \hat{s}$ 
 and $\Delta \hat{e}$, 
\begin{equation}
\{\Delta \hat{s}, \Delta \hat{e}\} = \phi_{r}([I_s;I_e])
\end{equation}
then, the refined proposal can be obtained:
\begin{equation}
\hat{s}_r = s_p - \Delta\hat{s}w_p, \qquad \hat{e}_r=e_p - \Delta\hat{e}w_p
\end{equation}
where $w_p = e_p - s_p$ is the length of the proposal.

Then, the reliability-aware supervision for instance level boundary regression can be formulated as:
\begin{equation}
{\mathcal L}_{reg}=\frac{1}{N_p} \sum_{i=1}^{N_p}Smooth{L1}(\hat{r}_{comp},1)
\end{equation}

where  $\hat{r}_{comp}$ represents the IoU between the refined proposal and the most reliable {proposal} ($\mathbf{R}\mathbf{P}$) that matches it.

Finally, the reliability-aware instance-level completeness learning has an overall objective function that consists of both regression and score losses weighted by a parameter $\lambda_2$, formulated as:
\begin{equation}
\mathcal{L}_{instance} = \mathcal{L}_{score} + \lambda_2\mathcal{L}_{reg}
\end{equation}

\begin{table*}[t]
\centering
\resizebox{\textwidth}{!}{
\begin{tabular}{c|l|ccccccc|ccc}
\hline\hline

\multirow{2}{*}{Supervision}       & \multicolumn{1}{c|}{\multirow{2}{*}{Method}} &
\multicolumn{7}{c|}{mAP@IoU (\%)}  & AVG  & AVG  & AVG  \\
    & \multicolumn{1}{c|}{}  & 0.1  & 0.2  & 0.3  & 0.4  & 0.5  & 0.6  & 0.7 & (0.1:0.5)  & (0.3:0.7) & (0.1:0.7)    \\
    \hline\hline
    
\multirow{4}{*}{\begin{tabular}{c}Frame-level\\(Full)\end{tabular}}
    & P-GCN \small {(ICCV'19)}  & 69.5  & 67.8  & 63.6  & 57.8  & 49.1  & -  & -  & 61.6  & - & - \\
    & TCANet \small {(CVPR'21)}  & -  & -  & 60.6  & 53.2  & 44.6  & 36.8  & 26.7  & -  & 44.3 & -\\
    & AFSD \small {(CVPR'21)} & -  & -  & \textbf{67.3}  & \textbf{62.4}  & \textbf{55.5}  & \textbf{43.7}  & \textbf{31.1}  & -  & \textbf{52.0} & -\\
    \hline\hline
    
\multirow{11}{*}{\begin{tabular}{c}Video-level\\(Weak)\end{tabular}}
    & CoLA \small {(CVPR'21)}   & 66.2  & 59.5  & 51.5  & 41.9  & 32.2  & 22.0  & 13.1  & 50.3  & 32.1 & 40.9  \\
    & TS-PCA \small {(CVPR'21)}  & 67.6  & 61.1  & 53.4  & 43.4  & 34.3  & 24.7  & 13.7  & 52.0  & 33.9 & 42.6\\
    & UGCT \small {(CVPR'21)}    & 69.2  & 62.9  & 55.5  & 46.5  & 35.9  & 23.8  & 11.4  & 54.0  & 34.6 & 43.6\\
    & FAC-Net \small {(ICCV'21)} & 67.6  & 62.1  & 52.6  & 44.3  & 33.4  & 22.5  & 12.7  & 52.0  & 33.1 & 42.2 \\
    & ACG-Net \small{(AAAI'22)} & 68.1  & 62.6  & 53.1  & 44.6  & 34.7  & 22.6  & 12.0  & 52.6  & 33.4 & 42.5\\
    & RSKP \small {(CVPR'22)}   & 71.3  & 65.3  & 55.8  & 47.5  & 38.2  & 25.4  & 12.5  & 55.6  & 35.9  & 45.1\\
    & DELU \small {(ECCV'22)}   & 71.5  & 66.2  & 56.5  & 47.7  & 40.5  & 27.2  & \textbf{15.3}  & 56.5  & 37.4  & 46.4\\
    & Li~\etal~ \small {(CVPR'23)}   & -  & -  & 56.2  & 47.8  & 39.3  & \textbf{27.5}  & 15.2  & -  & 37.2  & -\\
    & Zhou~\etal \small {(CVPR'23)}   & \textbf{74.0}  & \textbf{69.4}  & \textbf{60.7}  & \textbf{51.8}  & \textbf{42.7}  & 26.2  & 13.1  & \textbf{59.7}  & \textbf{38.9}  & \textbf{48.3}\\

    \hline
    \multirow{5}{*}{\begin{tabular}{c}Point-level\\(Weak)\end{tabular}}
    & SF-Net \small {(ECCV'20)}  & 68.3  & 62.3  & 52.8  & 42.2  & 30.5  & 20.6  & 12.0   & 51.2  & 31.6  & 41.2\\
    & Ju~\etal~ \small {(ICCV'21)}  & 72.3  & 64.7  & 58.2  & 47.1  & 35.9  & 23.0  & 12.8   & 55.6  & 35.4  & 44.9\\
    & LACP \small {(ICCV'21)}  & 75.7  & 71.4  & 64.6  & 56.5  & 45.3  & 34.5  &  21.8  & 62.7  &  44.5  & 52.8\\
    & CRRC-Net \small{(TIP'22)} & 77.8  & 73.5  & 67.1  & 57.9  & 46.6  & 33.7  &  19.8  & 64.6  &  45.1  & 53.8\\
    &  \textbf{\modelname~(Ours)}  & \textbf{85.6}  & \textbf{81.6}  & \textbf{74.3}  & \textbf{64.3}  & \textbf{52.2}  & \textbf{39.8}  & \textbf{24.8}  & \textbf{71.6}{$^{\uparrow7.0}$}  &  \textbf{51.1}{$^{\uparrow6.0}$} &\textbf{60.3}{$^{\uparrow6.5}$} \\
    
    \hline\hline
    
\end{tabular}
}
\caption{
Comparisons of detection performance on THUMOS14.
We include the methods under video-level and frame-level supervision for reference.
We utilize the same annotations under the point-level supervision as ~\cite{lee2021learning}. $\uparrow$ denotes the relative performance gain between our method (the best) and the second-best method under point-level supervision setting.}
\label{table:thumos_benchmark}
\end{table*}

\begin{table*}[!t]
\centering
\resizebox{\textwidth}{!}{
\begin{tabular}{l|ccccc|ccccc|cccc}
\hline\hline
 & \multicolumn{5}{c|}{GTEA} & \multicolumn{5}{c|}{BEOID} & \multicolumn{4}{c}{ActivityNet1.3} \\
\hline
\multirow{2}{*}{Method}& \multicolumn{5}{c|}{mAP@IoU (\%)} & \multicolumn{5}{c|}{mAP@IoU (\%)} & \multicolumn{4}{c}{mAP@IoU (\%)} \\
\multicolumn{1}{c|}{} & 0.1  & 0.3  & 0.5  & 0.7 &AVG[0.1:0.7] & 0.1  & 0.3  & 0.5  & 0.7 &AVG[0.1:0.7] & 0.5  & 0.75  & 0.95 &AVG[0.5:0.95] \\
\hline \hline
SF-Net \small(ECCV'20) & 58.0  & 37.9  & 19.3  & 11.9  & 31.0 & 62.9  & 40.6  & 16.7  & 3.5  & 30.9 & - & - & - & - \\
Ju~\etal~ \small(ICCV'21)  & 59.7  & 38.3  & 21.9  & 18.1  & 33.7 & 63.2  & 46.8  & 20.9  & 5.8  & 34.9 & - & - & - & - \\
Li~\etal~ \small(CVPR'21)  & 60.2  & 44.7  & 28.8  & 12.2  & 36.4 & 71.5  & 40.3  & 20.3  & 5.5  & 34.4 & - & - & - & - \\
LACP \small(ICCV'21) & 63.9  & 55.7  & 33.9  & \textbf{20.8}  & 43.5  & 76.9  & 61.4  & 42.7  & 25.1  & 51.8 & 40.4  & 24.6  & 5.7 & 25.1 \\
CRRC-Net \small(TIP'22)  & -  & -  & - & - & - & -  & -  & - & - & - & 39.8  & 24.1  & 5.9 & 24.0 \\
\textbf{\modelname~(Ours)}  & \textbf{72.6}  & \textbf{61.1}  & \textbf{37.3}  & 17.5  & \textbf{47.3{$^{\uparrow3.8}$}}  
& \textbf{78.5}  & \textbf{72.1}  & \textbf{55.3}  & \textbf{26.1}  & \textbf{59.4{$^{\uparrow7.6}$}}
& \textbf{42.8}  & \textbf{27.2}  & \textbf{8.0} & \textbf{27.1{$^{\uparrow2.0}$}} \\
\hline \hline

\end{tabular}
}
\caption{
Comparisons of detection performance on GTEA, BEOID, and ActivityNet1.3 datasets.
}
\label{table:multi_benchmark}
\end{table*}

\subsection{Temporal Action Localization Inference} 

We first extract the snippet-level prediction of predicted class $P_c$ and class-agnostic attention $A$ of each video, which is used to generate candidate proposals,represented as $(s_i,e_i,c_i,p_{OIC})$.
Then, we input the instance-level feature of each proposal to the Score and Regression heads, which derives two parts of predicted proposals: the score refined part $(s_i,e_i,c_i,p_{OIC}+\hat{p}_{comp})$ and the boundary refined part $(\hat{s_r},\hat{e}_r,c_i,p_{OIC}+\hat{p}_r)$, $\hat{p}_r$ is the completeness score of refined proposal estimated by the trained Score Head. Finally, we combine them and employ class-wise soft-NMS~\cite{bodla2017soft} to remove duplicate proposals.

\begin{table}[!t]
\centering
\resizebox{0.47\textwidth}{!}{
\begin{tabular}{ccc|cccc|c}
    \hline\hline
    \multicolumn{3}{c|}{{Snippet-level}} 
    &\multicolumn{4}{c|}{{Instance-level}}
    & \textbf{AVG} \\
    \hline
    $\mathcal{L}_{base}$ & $\mathcal{L}_{contra}$  & RAB  
    &$\mathcal{L}_{reg}$  & $\mathcal{L}_{score}$ &RP &NP  & [0.1:0.7] \\
    \hline
    $\checkmark$ &   &   &   &   &   &   & 49.4  \\
    $\checkmark$ &$\checkmark$  &   &   &   &   &   &51.5  \\
    $\checkmark$ &$\checkmark$   &$\checkmark$   &   &   &   &   &54.7{$^{\uparrow5.3}$}   \\
    \hline
    $\checkmark$ &$\checkmark$   &$\checkmark$   &$\checkmark$   &   &   &   & 56.1  \\
    $\checkmark$ &$\checkmark$   &$\checkmark$   &   &$\checkmark$   &   &   & 57.1  \\
    $\checkmark$ &$\checkmark$   &$\checkmark$   &$\checkmark$   &$\checkmark$   &   &   &57.8   \\
    $\checkmark$ &$\checkmark$   &$\checkmark$   &$\checkmark$   &   &$\checkmark$   &   &56.8   \\
    $\checkmark$ &$\checkmark$   &$\checkmark$   &   &$\checkmark$   &$\checkmark$   &   &58.1   \\
    $\checkmark$ &$\checkmark$   &$\checkmark$   &$\checkmark$   & $\checkmark$  & $\checkmark$  &   &59.1  \\
    $\checkmark$ &$\checkmark$   &$\checkmark$   &$\checkmark$   & $\checkmark$  & $\checkmark$  &$\checkmark$   & \textbf{60.3}{$^{\uparrow10.9}$} \\
    \hline
    $\checkmark$ &   &   &$\checkmark$   &$\checkmark$   &$\checkmark$   &$\checkmark$   & 53.9  \\
    $\checkmark$ & $\checkmark$   &   &$\checkmark$   &$\checkmark$   &$\checkmark$   &$\checkmark$   & 56.2   \\
    $\checkmark$ & $\checkmark$   &$\checkmark$    &$\checkmark$   &$\checkmark$   &$\checkmark$   &$\checkmark$   & \textbf{60.3}{$^{\uparrow10.9}$}   \\
    \hline\hline
\end{tabular}
}
\caption{
Ablation study on THUMOS14.
$\uparrow$ denotes the relative gain between each setting and baseline ($\mathcal{L}_{base}$ only).
}
\label{table:ablation_studies}
\end{table}

\section{Experiments}
\begin{figure*}[t]

\centering
\includegraphics[scale=0.265]{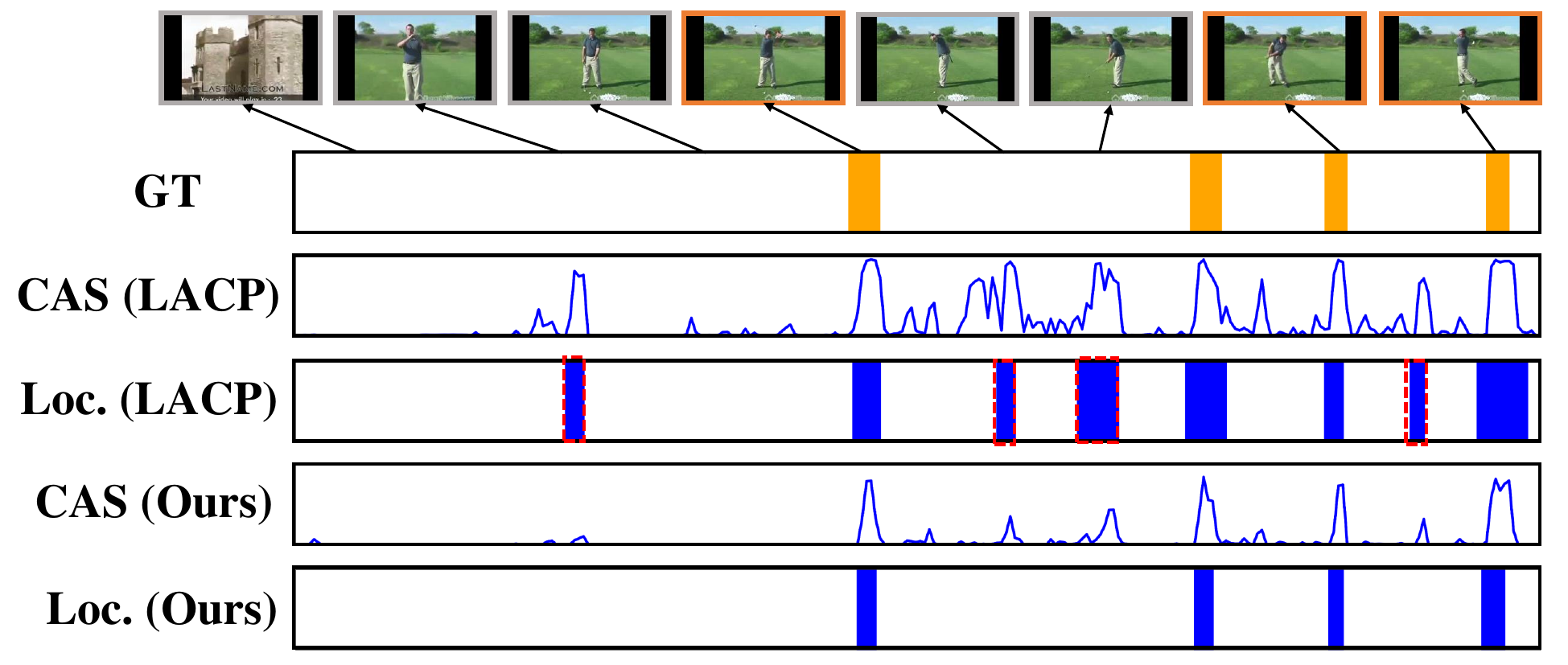}
\centering
\includegraphics[scale=0.265]{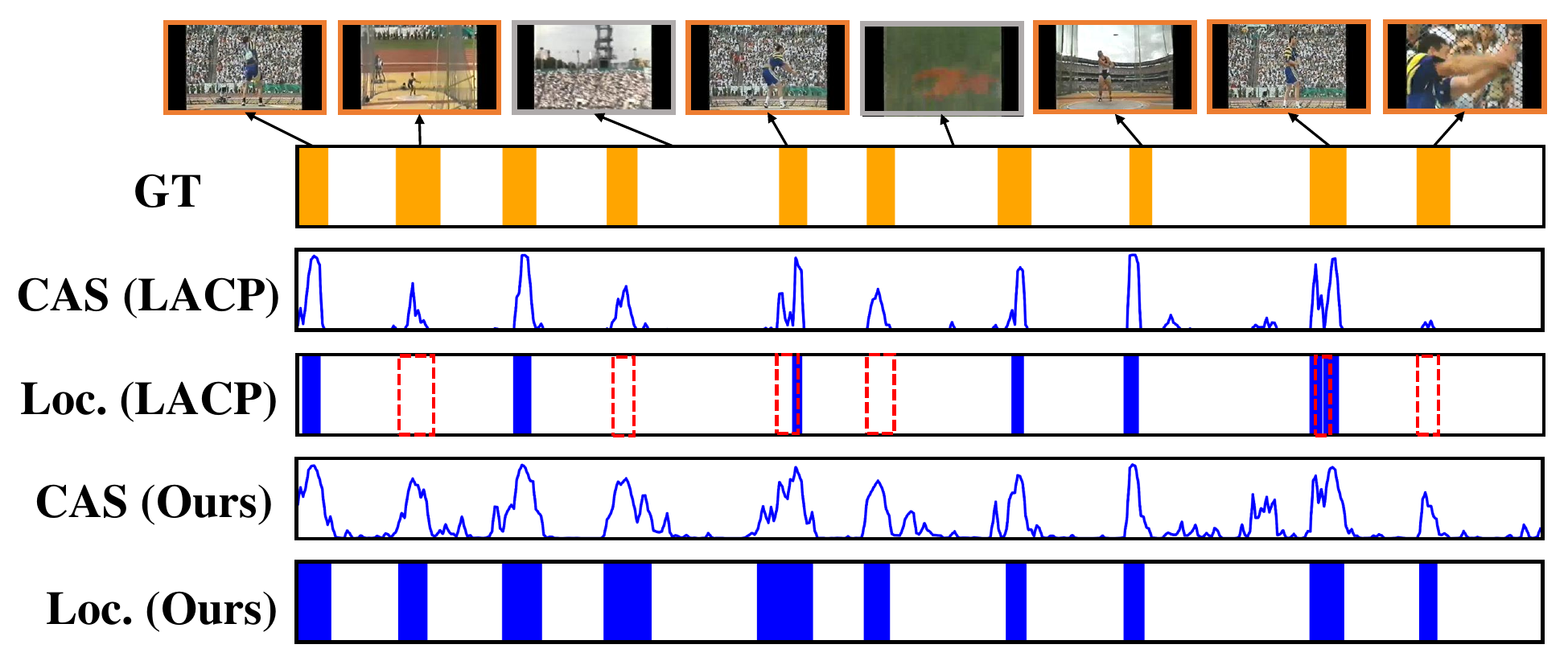} 
\caption{Qualitative results for two action categories, \textit{GolfSwing} (left) and \textit{HammerThrow} (right), on THUMOS14. We compare the detection results of \modelname~and LACP. The orange and blue bars indicate the ground truth and predicted localization results, respectively; Blue curves represent snippet-level prediction. Prediction errors are bound with red bounding boxes. }
\label{quality_result}
\end{figure*}

\begin{figure}[t]
\centering 
\includegraphics[scale=0.24]{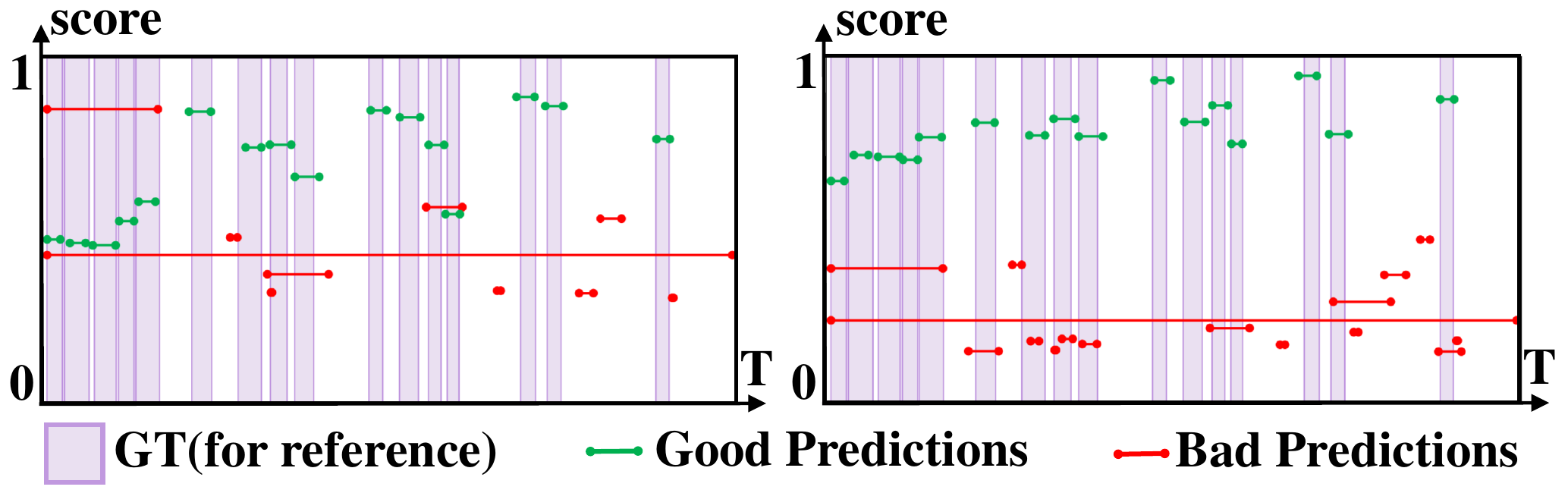}    
\caption{Visualization of detection results on THUMOS14 dataset before (left) and after (right) instance-level completeness learning. 
The x-axis and y-axis represent time and the reliability score, respectively. 
We observe that the discrepancy between good and bad predictions is enlarged significantly after instance-level completeness learning.
}
\label{fig:completeness}
\end{figure}
\subsection{Experimental Setup}

\noindent\textbf{Datasets}. We conduct our experiments on four popular action localization datasets, with only point-level annotations used for training.
{In our experiments, we utilize the point-level annotations provided in~\cite{lee2021learning} for fair comparison.}
(1) \textbf{THUMOS14}~\cite{thumos} provides 413 untrimmed sports videos for 20 action categories, including 200 videos for training and 213 videos for testing,
and each video contains an average of 15 action instances. 
Action instance lengths and video lengths vary widely, which makes this dataset challenging.
(2) \textbf{GTEA}~\cite{gtea} provides 28 videos of 7 fine-grained daily activities in a kitchen. Four different subjects perform an activity, and each video contains about 1,800 frames.
(3) \textbf{BEOID}~\cite{beiod} provides 58 video samples with 30 action classes with an average duration of 60$s$. There is an average of 12.5 action instances per video.
(4) \textbf{ActivityNet 1.3}~\cite{caba2015activitynet} provides 10,024 training, 4,926 validation, and 5,044 test videos with 200 action classes. Each video includes 1.6 action instances on average.

\noindent\textbf{Evaluation metric}.
We follow the standard protocols to evaluate with mean Average Precision (mAP) under different intersection over union (IoU) thresholds. A proposal is regarded as positive only if both IoU exceeds the set threshold and the category prediction is correct.

\noindent\textbf{Implementation Details.}
For a fair comparison, we follow existing method~\cite{lee2021learning} to divide each video into 16-frame snippets and use two-stream I3D network pre-trained on Kinetics-400~\cite{i3d} as the feature extractor.
For THUMOS14,
we use the Adam optimizer with a learning rate of 1e-4 and a weight decay of 1e-3, and the batch size is set to 16. The hyper-parameters are set by grid search: $\tau=0.1$, $\mu =0.999$, $\lambda_1=\lambda_2=1$.
The video-level threshold is set to 0.5, the $\theta_{P}$ spans from 0 to 0.25 with a step size of 0.05, the $\theta_{A}$ spans from 0 to 0.1 with a step size of 0.01.
The number of RAB is set to 2.

\subsection{Comparison with State-of-The-Art Methods}
We evaluate the effectiveness of our proposed method by comparing it against the most recent fully-supervised and weakly-supervised temporal action localization methods.

\noindent\textbf{THUMOS14.} Our proposed method, \textit{i.e.}, \modelname, achieves state-of-the-art performance on THUMOS14 testing set for point-level weakly-supervised temporal action localization. Compared to previous state-of-the-art methods in Table~\ref{table:thumos_benchmark}, \modelname~has an average mAP of 60.3\% for IoU thresholds of 0.1:0.7, outperforming the prior SoTA method~\cite{fu2022compact} by 6.5\% for the same thresholds. Notably, our point-supervised method is able to achieve comparable performance with competitive fully-supervised methods, such as AFSD (51.1\% vs. 52.0\% in average mAP for IoU thresholds of 0.3:0.7). Moreover, \modelname~demonstrates superior detection performance compared to video-level weakly-supervised methods with similar labeling cost, thanks to the position information provided by point labels.

\noindent\textbf{GTEA \& BEOID \& ActivityNet 1.3.} We demonstrate the generality and superiority of \modelname~ on diverse benchmarks in Table~\ref{table:multi_benchmark}. Our method significantly outperforms existing methods, achieving improvements of  3.8\%, 7.6\%, and {2.0\%}, on GTEA, BEOID, and ActivityNet 1.3, respectively.

\subsection{Ablation Study}
To further analyze the contribution of model components compared to the baseline setting (with a detection result of 49.4\%), we perform a set of ablation studies on THUMOS14. The results are summarized in Table~\ref{table:ablation_studies}.

\noindent\textbf{Snippet-level Discrimination Learning.} The introduction of contrastive loss increases performance by 2.1\%.
Contrastive optimization not only reduces classification errors but also improves the model's ability to distinguish between action and background, thereby improving detection performance.
The introduction of Reliability-aware Attention Block (RAB) further improves detection performance by 3.2\%.
We speculate that the introduction of RAB increases the attention on less reliable action snippets, thus detecting more non-discriminative actions.

\noindent\textbf{Instance-level Completeness Learning.} We see the introduction of regression loss and score loss significantly increases the detection performance. 
The introduction of reliable proposals and negative proposals generated based on point annotations further boosts the results.
These results demonstrate that the components of instance-level completeness learning complement each other and make the network estimate the proposal score and boundaries more accurately.

\subsection{Qualitative Results}
\noindent\textbf{Qualitative Comparison.} In Fig.~\ref{quality_result}, we compare our \modelname~with LACP for temporal action localization on test videos in THUMOS14. Our model shows more accurate detection of action instances. 
In specific, for \textit{GolfSwing} action, our method effectively distinguishes between action and background snippets, mitigating false action predictions that LACP struggles with;
for \textit{HammerThrow} action, our method detects more complete snippets than LACP, which has lower activation values on non-discriminative action snippets.

\noindent\textbf{Effect of Instance-level Completeness Learning.} Fig.~\ref{fig:completeness} shows that completeness learning helps our method reduce the score of overcomplete and false positive proposals, leading to improved detection results.

\section{Conclusion}

{
This paper introduces a new framework named HR-Pro for point-supervised temporal action localization.
HR-Pro comprises two reliability-aware stages that efficiently propagate high-confidence cues from point annotations at both the snippet and instance level, which enables the network to learn more discriminative snippet representation and more reliable proposals.
Extensive experiments on multiple benchmarks demonstrate that HR-Pro significantly outperforms existing methods and achieves state-of-the-art results, which demonstrates the effectiveness of our method and the potential of point annotations.
}

\section*{Acknowledgement}
This work is supported by the National Natural Science Foundation of China under grant U22B2053.

\bibliography{aaai24}

\clearpage
\setcounter{table}{0}  
\setcounter{figure}{0}
\setcounter{equation}{0} 
\renewcommand{\thetable}{\Alph{table}}
\renewcommand{\thefigure}{\Alph{figure}}

\section{Appendix}

\subsection{More Implementation Details}
\textbf{Pseudo Snippets Generation.}
As discussed in the main paper, we use the class-agnostic attention sequence output by the model and ground truth point annotations to generate pseudo snippets to supervise snippet-level predictions. The pseudo-code of the pseudo snippets generation algorithm is described in the Algorithm~\ref{alg:psg}.
In practice, snippet action threshold $\theta_{1}$ is set to 0.95, and snippet background threshold $\theta_{2}$ is set to 0.1.

\noindent\textbf{Feature Embedding.}
The detail of feature embedding layer $\phi_e$ can be defined as:
\begin{equation}
\mathbf{X}^e=\phi_e(\mathbf{X}, \mathbf{w}_e)
\label{eq_1}
\end{equation}
Here, 
$\mathbf{X}^e\in{\mathbb R}^{T\times D}$ represents the embedded features, 
$\mathbf{w}_e$ denotes learnable parameters.

\noindent\textbf{Genaration of $P$ and $A$.}
The detailed process of generation of $P$ and $A$ can be formulated as follows:
\begin{equation}
\mathbf{S}=\sigma({\mathbf{X}^e}{\mathbf{w}_{cls}}+\mathbf{b}_{cls}), \quad
\mathbf{A}=\sigma({\mathbf{X}^e}{\mathbf{w}_a}+\mathbf{b}_a)
\label{eq_2}
\end{equation}
where $ \mathbf{w}_{cls}\in  {\mathbb R}^{D \times C} $ and $ \mathbf{w}_{a}\in  {\mathbb R}^{D \times 1} $ are the learnable parameters of snippet-level classifier and temporal attention mechanisms, $ \mathbf{b}_{cls} $ and $ \mathbf{b}_{a} $  are the corresponding biases, and $ \sigma(\cdot) $ represents the \textit{sigmoid} activation function.

\noindent\textbf{Details of $\phi_s$ and $\phi_{r}$}
$\phi_s$ and $\phi_{r}$ are implemented with a two-layer Conv-lD block with an activation layer.

\noindent\textbf{Hardware and Software Settings.}  An NVIDIA Linux workstation (GPU: 1$\times$ 2080Ti) is used for our experiments. Our codebase is built on PyTorch 1.7.0.

\begin{algorithm}[tb]
\caption{Pseudo Snippets Generation.}
\label{alg:psg}
\raggedright
\textbf{Input}: Class-agnostic attention sequence $\mathbf{A}\in{\mathbb R}^T$, ground truth point annotations $\mathbf{G}=\{(p_i,y_{p_i})\}_i^N$.\\
\textbf{Output}: Pseudo action snippets $\mathcal{T}^+=\{t_i\}_i^{N_{act}}$, pseudo background snippets $\mathcal{T}^-=\{t_j\}_j^{N_{bkg}}$.

\begin{algorithmic}[1] 
\STATE Let $\mathcal{T}^+ \gets \varnothing$, $\mathcal{T}^- \gets \varnothing$.

\FOR{every $p_i \in \mathbf{G}$}
    \STATE \textbf{if} $i=0$, \textbf{then} $p_{i-1} \gets 0$
    \STATE \textbf{end if}
    \STATE \textbf{if} $i=N$, \textbf{then} $p_{i+1} \gets T$
    \STATE \textbf{end if}

    \FOR{$t = p_i$ {\bfseries to} $p_{i-1}$}
        \STATE \textbf{if} $\mathbf{A}[t] > \theta_{1}$, \textbf{then} $\mathcal{T}^+ \gets t \cup \mathcal{T}^+$, \textbf{else break}
        \STATE \textbf{end if}

        \STATE \textbf{if} $\mathbf{A}[t] < \theta_{2}$ \textbf{or} $t=argmin(A[p_i-1:p_i])$, \textbf{then} $\mathcal{T}^- \gets t \cup \mathcal{T}^-$
        \STATE \textbf{end if}
    \ENDFOR

    \FOR{$t = p_i$ {\bfseries to} $p_{i+1}$}
        \STATE \textbf{if} $\mathbf{A}[t] > \theta_{1}$, \textbf{then} $\mathcal{T}^+ \gets t \cup \mathcal{T}^+$, \textbf{else break}
        \STATE \textbf{end if}

        \STATE \textbf{if} $\mathbf{A}[t] < \theta_{2}$ \textbf{or} $t=argmin(A[p_i:p_{i+1}])$, \textbf{then} $\mathcal{T}^- \gets t \cup \mathcal{T}^-$
        \STATE \textbf{end if}
    \ENDFOR
\ENDFOR
\STATE \textbf{Return} $ \mathcal{T}^+, \mathcal{T}^-$
\end{algorithmic}
\end{algorithm}
\subsection{Additional Experiments}
We present additional experiments to further demonstrate the effectiveness and robustness of our proposed HR-Pro.

\textit{\textbf{Is HR-Pro robust to the distribution of point annotations?}}
Yes, as mentioned in Table 1, we utilize the same point-level annotations in~\cite{lee2021learning} for a fair comparison, which randomly selects a timestamp within the specified Gaussian distribution between the start and end of each action instance as the point annotation.
To further illustrate the influence of point annotation selection strategies, we supplement the ablation studies on THUMOS14 in Table~\ref{table:ablation_studies_point} to ablate different distributions (e.g., \textit{uniform}, \textit{center}, \textit{gaussian}, \textit{center} indicates the center timestamp of each action instance is selected as the point annotation.).
We observe that our method exhibits slightly worse performance in the \textit{uniform} distribution compared to the other distributions. 
We think this is because less discriminative points have a higher likelihood of being annotated.
As a result, the low confidence of their surrounding snippets leads to a decrease in the number of pseudo-action samples.
The results on the \textit{gaussian} and \textit{center} distributions are close, which demonstrates the robustness of our method to point distributions.
\begin{table}[!t]
\centering
\resizebox{0.48\textwidth}{!}{
\begin{tabular}{c|cccc|c}
    \hline\hline
    \multirow{2}{*}{Point Distribution} & \multicolumn{4}{c|}{mAP@IoU (\%)}  & \textbf{AVG} \\
      & 0.1  & 0.3  & 0.5  & 0.7 & [0.1:0.7] \\
    \hline
      {Uniform} & 84.4 & 72.8 & 48.6 & 21.6 & 58.8 \\
      {Center} & 84.8 & 74.2 & 53.4 & 26.9 & 60.8 \\
      {Gaussian (default)} & 85.6 & 74.3 & 52.2 & 24.8 & 60.3 \\
    \hline\hline
\end{tabular}
}
\vspace{-2mm}
\caption{Comparison of the point-level annotations from different distributions on THUMOS14.}
\label{table:ablation_studies_point}
\vspace{-6mm}
\end{table}

\vspace{-2mm}
\subsection{More Qualitative Results}
\begin{figure*}[t]

\centering
\includegraphics[scale=0.53]{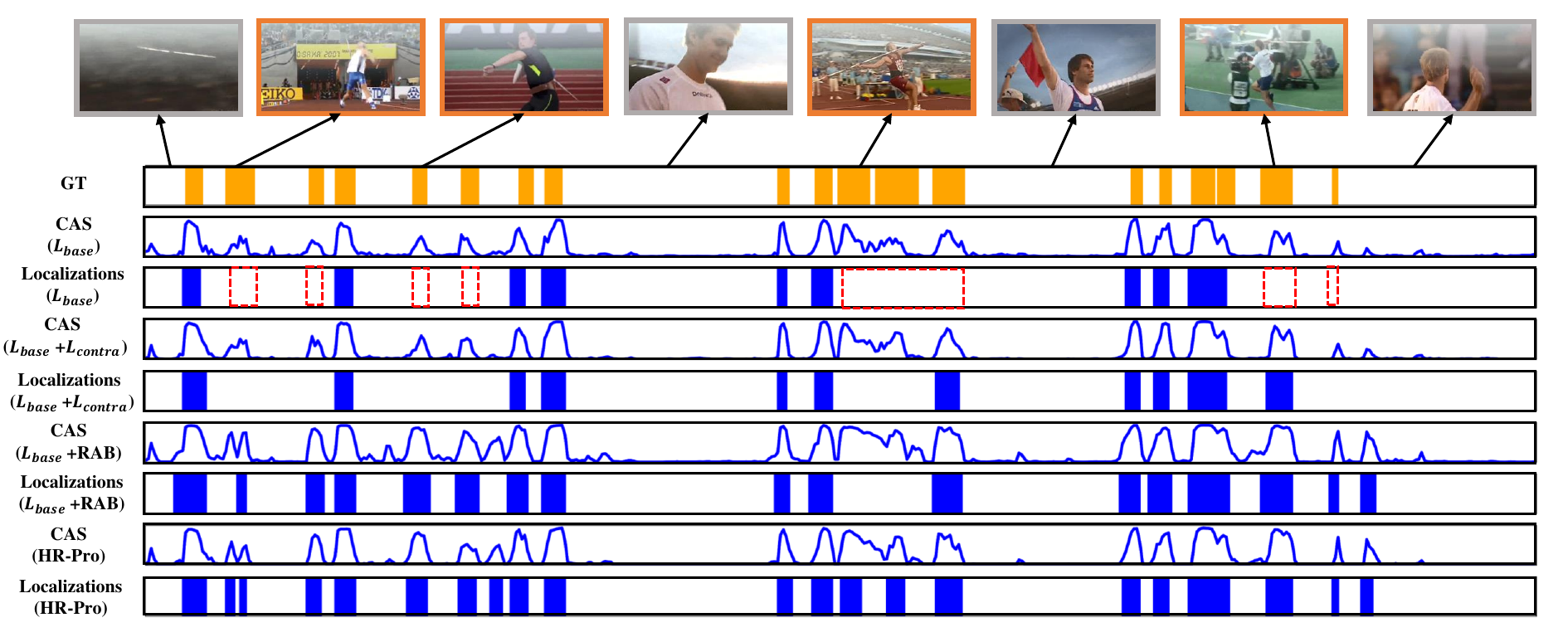}
\centering
\includegraphics[scale=0.53]{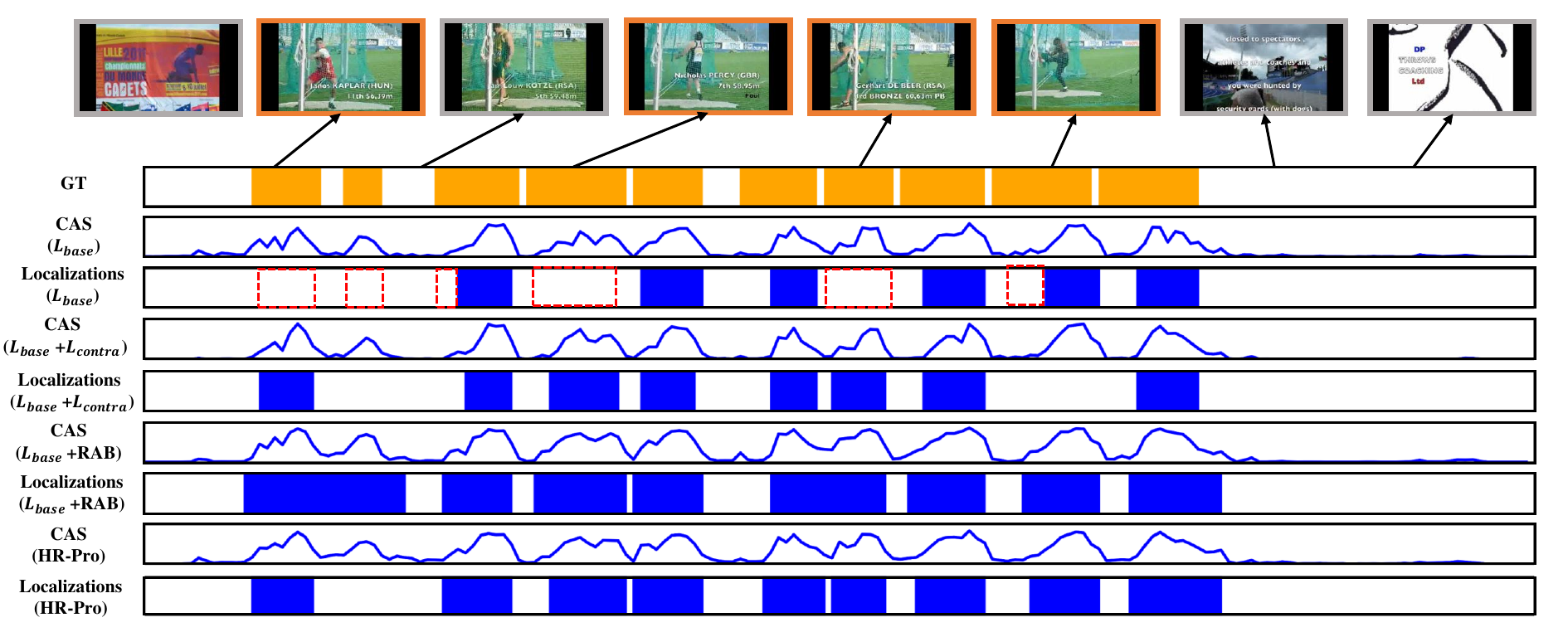}
 \vspace{-4mm}
\caption{Qualitative Ablation results on THUMOS14. We show examples of two different action categories: \textit{JavelinThrow} (top) and \textit{ThrowDiscus} (bottom). We compare the detection results of different
types for the combination of the designed components for each video. The orange bar and blue bar respectively represent ground truths and localization results, and the blue curve represents snippet-level prediction. Prediction errors are highlighted with red bounding boxes.}
\label{quality_result_supply}
\end{figure*}

\noindent\textbf{Qualitative Ablation}
We qualitatively compare our method with the baseline model on the THUMOS14 dataset, we present the visualization results of different types for the combination of the designed components for each video example, including snippet-level predictions output by the model and localization results, as shown in Figure~\ref{quality_result_supply}.

\begin{figure}[t]
\centering
\subfigure[\textbf{Baseline}]{
	\includegraphics[scale=0.325]{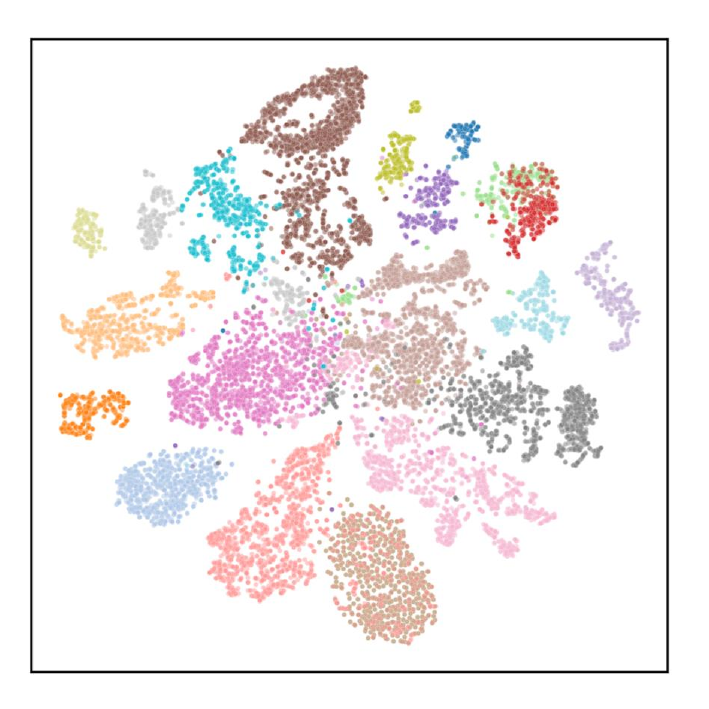}
        }      
\subfigure[\textbf{HR-Pro}]{
	\includegraphics[scale=0.325]{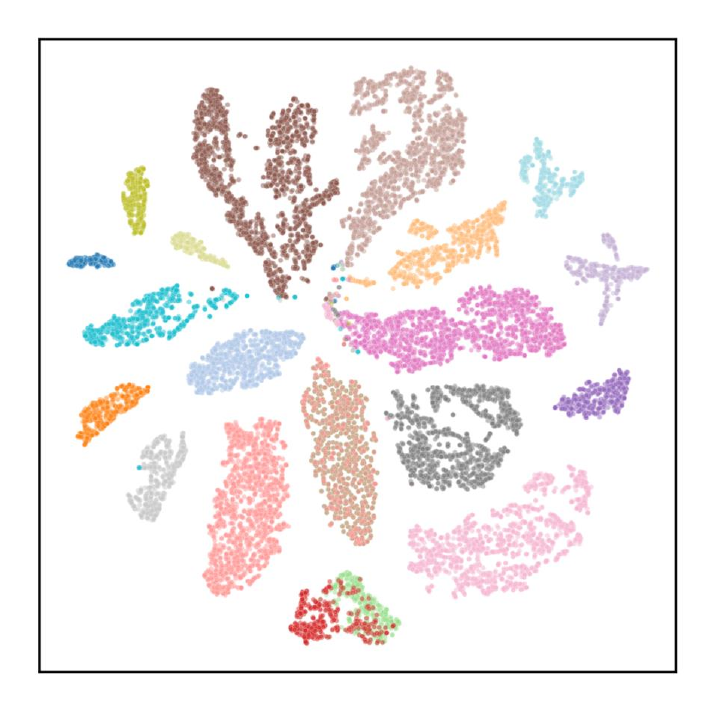}
        }  
\vspace{-4mm}
\caption{t-SNE visualization of feature space on THUMOS14 dataset. 
a) and b) show  the  results of the output feature of baseline and HR-Pro, respectively, where dots represent snippet features and different colors indicate different action categories.}
\label{tSNE}
\end{figure}

\noindent\textbf{Feature Embedding Analysis}
We use t-SNE~\cite{tsne} to visualize the feature embedding in Figure~\ref{tSNE}.
We find that the action features of different categories in the baseline method were relatively scattered.
Through reliability-aware contrastive loss, the network pushes together the action features of the same class and pulls away the action features of different classes, thus improving the model's ability to distinguish between different classes, making the features of each class more centralized and reducing classification errors.

\begin{figure*}[t]
\centering
\includegraphics[scale=0.4]{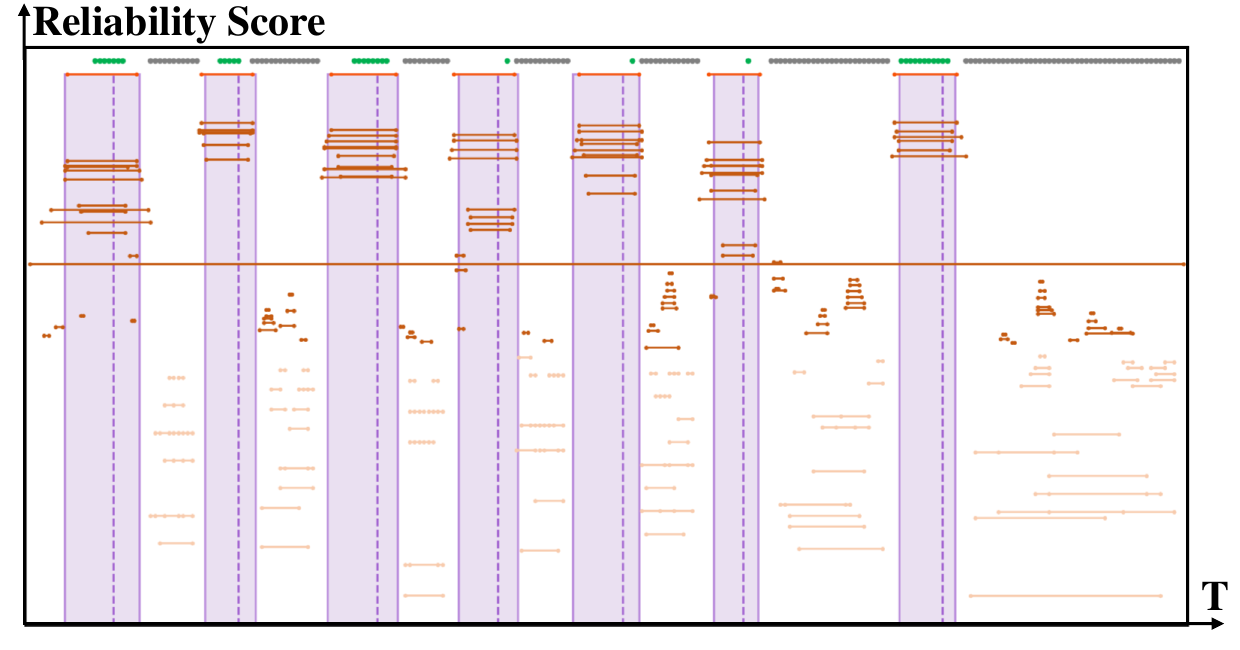}
\centering
\includegraphics[scale=0.4]{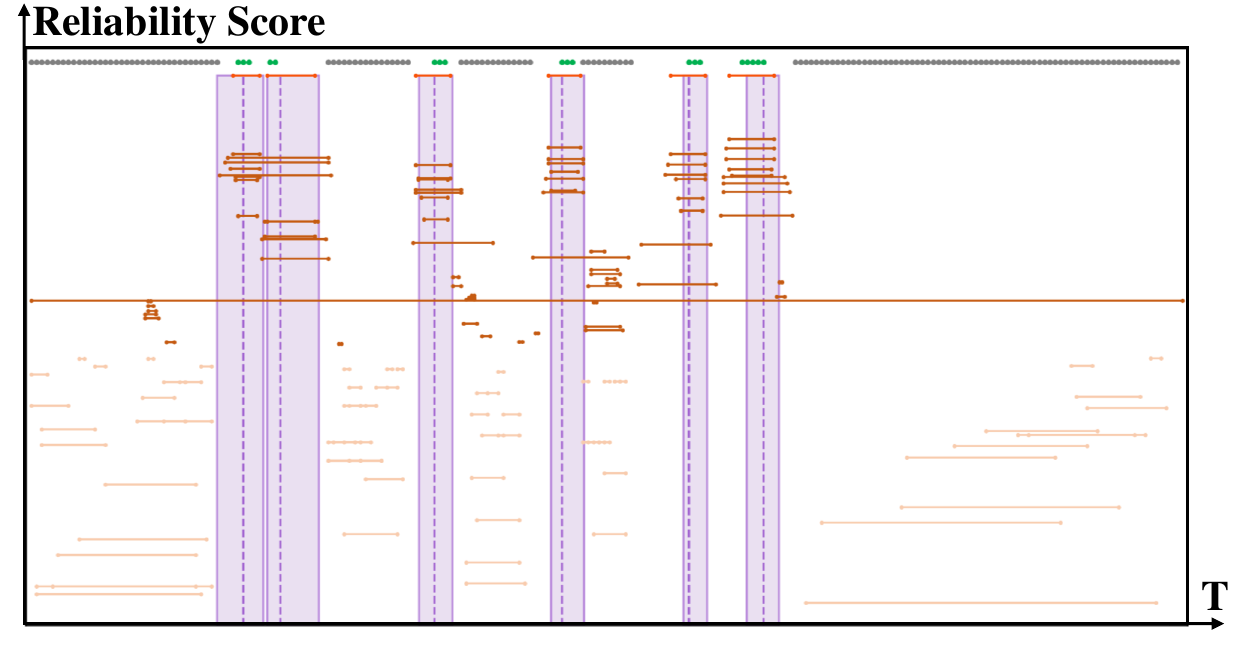}
\centering
\includegraphics[scale=0.4]{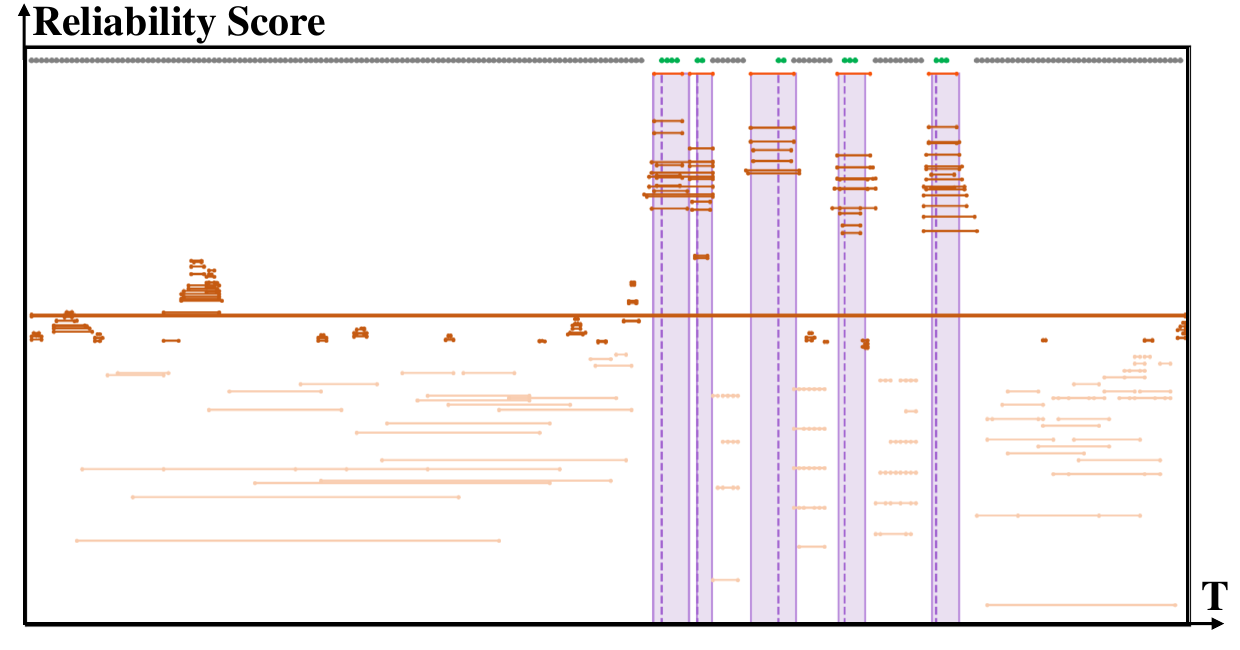}
\centering
\includegraphics[scale=0.4]{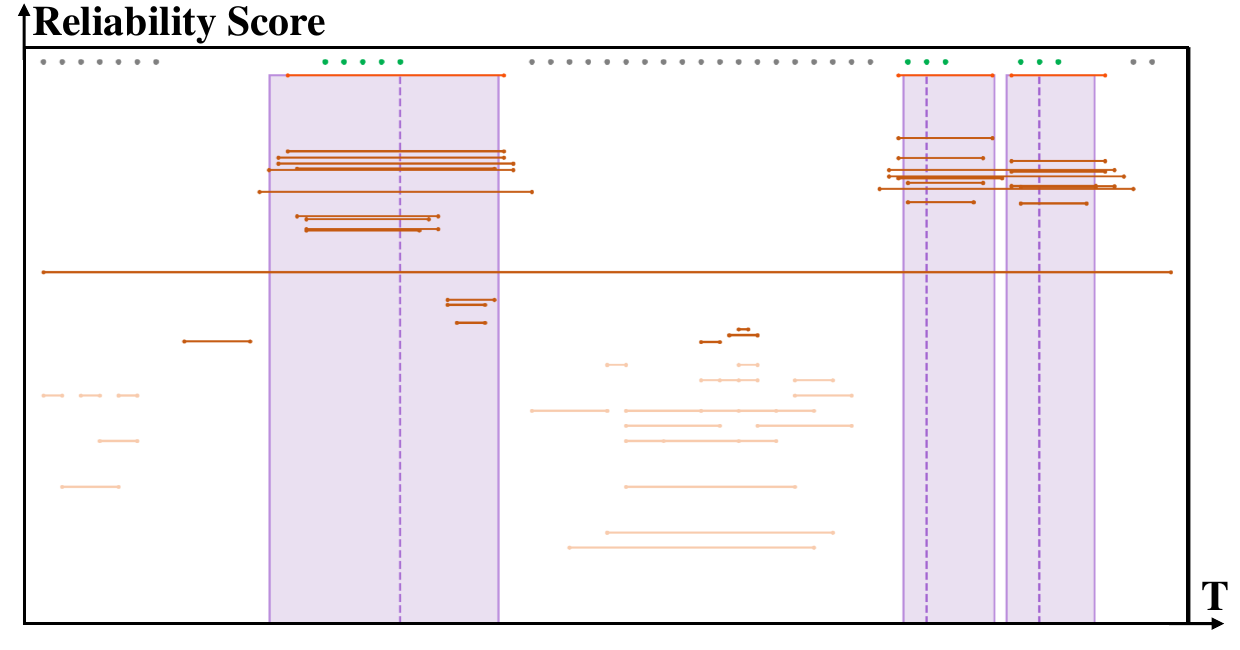}
 
\caption{Reliability Visualization results on THUMOS14 (zoom in for observation). We show examples of 4 different videos. The green and gray dots at the top of the figure respectively represent pseudo action snippets and pseudo background snippets. Orange lines denote different types of proposals, with the darkest color representing Reliable Proposals (RP), followed by Positive Proposals (PP) and Negative Proposals (NP). The purple region and vertical line correspond to Ground Truth (GT) and point annotations.}
\label{quality_rv}
\end{figure*}

\noindent\textbf{Reliability Visualization}
We visualize the reliable snippets and instances during the training process, and compare them with the Ground Truth in Figure~\ref{quality_rv} . We observed that these reliable snippets and instances exhibit high quality thanks to the temporal prior provided by point annotations. Furthermore, following the propagation of reliability at the snippet and instance level, they furnish the network with more dense guidance.


\end{document}